\documentclass[conference]{IEEEtran}
\IEEEoverridecommandlockouts
\usepackage{cite}
\usepackage{amsmath,amssymb,amsfonts}
\usepackage{algorithmic}
\usepackage{graphicx}
\usepackage{textcomp}
\usepackage{xcolor}
\usepackage{subcaption}
\usepackage{booktabs}
\usepackage[normalem]{ulem}
\def\BibTeX{{\rm B\kern-.05em{\sc i\kern-.025em b}\kern-.08em
    T\kern-.1667em\lower.7ex\hbox{E}\kern-.125emX}}
\useunder{\uline}{\ul}{}
\begin{document}

\makeatletter
\newcommand{\linebreakand}{%
  \end{@IEEEauthorhalign}
  \hfill\mbox{}\par
  \mbox{}\hfill\begin{@IEEEauthorhalign}
}
\makeatother

\title{Supervised Radio Frequency Interference Detection with SNNs
\thanks{This work was supported by a Westpac Future Leaders Scholarship, an Australian Government Research Training Program Fees Offset and an Australian Government Research Training Program Stipend.}
}

\author{\IEEEauthorblockN{Nicholas J. Pritchard}
\IEEEauthorblockA{\textit{International Centre for Radio Astronomy Research} \\
\textit{University of Western Australia}\\
Perth, Australia \\
0000-0003-0587-2310}
\and
\IEEEauthorblockN{Andreas Wicenec}
\IEEEauthorblockA{\textit{International Centre for Radio Astronomy Research} \\
\textit{University of Western Australia}\\
Perth, Australia \\
0000-0002-1774-5653}
\linebreakand
\IEEEauthorblockN{Mohammed Bennamoun}
\IEEEauthorblockA{\textit{School of Physics, Mathematics and Computing} \\
\textit{University of Western Australia}\\
Perth, Australia \\
0000-0002-6603-3257}
\and
\IEEEauthorblockN{Richard Dodson}
\IEEEauthorblockA{\textit{International Centre for Radio Astronomy Research} \\
\textit{University of Western Australia}\\
Perth, Australia \\
0000-0003-0392-3604}
}

\maketitle

\begin{abstract}
Radio Frequency Interference (RFI) poses a significant challenge in radio astronomy, arising from terrestrial and celestial sources, disrupting observations conducted by radio telescopes. Addressing RFI involves intricate heuristic algorithms, manual examination, and, increasingly, machine learning methods. Given the dynamic and temporal nature of radio astronomy observations, Spiking Neural Networks (SNNs) emerge as a promising approach. In this study, we cast RFI detection as a supervised multi-variate time-series segmentation problem. Notably, our investigation explores the encoding of radio astronomy visibility data for SNN inference, considering six encoding schemes: rate, latency, delta-modulation, and three variations of the step-forward algorithm. We train a small two-layer fully connected SNN on simulated data derived from the Hydrogen Epoch of Reionization Array (HERA) telescope and perform extensive hyper-parameter optimization. Results reveal that latency encoding exhibits superior performance, achieving a per-pixel accuracy of 98.8\% and an f1-score of 0.761. Remarkably, these metrics approach those of contemporary RFI detection algorithms, notwithstanding the simplicity and compactness of our proposed network architecture. This study underscores the potential of RFI detection as a benchmark problem for SNN researchers, emphasizing the efficacy of SNNs in addressing complex time-series segmentation tasks in radio astronomy.
\end{abstract}

\begin{IEEEkeywords}
radio astronomy, spiking neural networks, neuromorphic computing, supervised learning, multi-variate time-series segmentation
\end{IEEEkeywords}

\section{Introduction}
Radio astronomy is a data-intensive science involving building and operating sophisticated observatories and high-performance computing facilities.
Since it is a fundamentally observational science, removing unwanted radio frequency interference (RFI) from artificial technology on Earth and in Space is crucial to an observatory's scientific capability.
Detecting RFI is a challenging task that increasingly requires adaptive data-driven methods in an era of increasing radio telescope sensitivity and the presence of terrestrial and satellite communication \cite{noauthor_report_2022}.
State-of-the-art RFI detection methods are either domain-specific traditional algorithms, e.g., \cite{offringa_aoflagger_2010} or CNNs, e.g., \cite{akeret_radio_2017}; both families of methods treat RFI detection as a boolean segmentation task on two-dimensional spectrograms (`visibility data' or `visibilities').
Traditional methods often require expert intervention, which becomes untenable as radio telescopes become larger, more sensitive, and produce more observations.
Inspired by SNNs' performance in audio-based processing tasks, we formulate RFI detection as a supervised time-series segmentation problem to explore how to encode visibility data for SNNs without prior offline analysis and exploit the time-varying nature of radio astronomy observations.
We present this task as a challenging benchmark problem ideal for SNNs and neuromorphic computing.

This work provides 1) a robust formulation of RFI detection as a time-series segmentation task that is appropriate for SNNs, 2) an investigation into encoding visibility data with latency, rate, delta-modulation, and three distinct variations on step-forward encodings for time-series segmentation, and 3) results for small SNNs trained and optimized with each encoding method on an RFI detection dataset.
The findings indicate that latency encoding is the most efficient encoding method.
This work establishes a baseline level of accuracy for SNNs trained for RFI detection without ANN conversion, introducing RFI detection as a challenging benchmark problem that the neuromorphic community should consider.
Thus, this research work is an essential contribution to RFI detection and demonstrates the potential of SNNs for solving critical problems in radio astronomy.
\section{Related Work}
Radio astronomy relies on spectrotemporal data processing, and while it is known that radio astronomy data is, in principle, a good match for SNNs and neuromorphic methods \cite{kasabov_evolving_2016}, there exist only a few attempts to apply SNNs anywhere in radio astronomy \cite{scott_evolving_2015}. Only one work applies SNNs to RFI detection \cite{pritchard_rfi_2024}.

RFI detection is a ubiquitous problem for radio telescopes, and as such, several methods exist to combat RFI contamination.
In the field, detecting RFI is called `flagging' as the output of a detection scheme is an array of boolean flags that can mask out contaminated pixels from downstream processing tasks.
Traditional methods to flag RFI build on variations of a cumulative sum algorithm, looking for regions of increased power in a spectrogram combined with domain and situation-specific additional filters and heuristics \cite{offringa_aoflagger_2010}.
While operational and efficient, these methods require constant adjustment, expert opinion, and, in the most challenging cases, resort to human labeling.

Applications of neural networks to RFI detection aim to ease the burden of manual labor.
They are primarily built around UNet-like CNNs \cite{akeret_radio_2017, yang_deep_2020}, unsupervised nearest-neighbor methods \cite{wolfaardt_machine_2016}, or auto-encoder-based anomaly detection routines based on finding nearest neighbors in latent space \cite{mesarcik_learning_2022}.
See \cite{dutoit_comparison_2024} for a recent comparison of deep learning approaches for RFI detection.
The only other application of SNNs to RFI detection employs ANN2SNN conversion on a pre-trained autoencoder to simplify the downstream RFI detection task \cite{pritchard_rfi_2024}.

This article focuses on a simulated dataset of observations from the Hydrogen Epoch of Reionization Array (HERA) in South Africa \cite{mesarcik_learning_2022-1}.
HERA's primary goal is to investigate the earliest periods of formation activity in the universe.
This simulated dataset helps characterize the type of RFI HERA will be subject to and has been benchmarked against traditional flagging algorithms, ANNs \cite{mesarcik_learning_2022, dutoit_comparison_2024}, and a converted SNN \cite{pritchard_rfi_2024}, therefore making this dataset an excellent initial target to investigate applying from-scratch trained SNNs to RFI detection.

Numerous well-known methods exist to encode continuous-valued time-series data, including well-known rate, latency, and delta-modulation methods \cite{yi_learning_2023}.
Encoding speech data, in particular, is greatly relevant to radio astronomy data since methods such as Speech2Spikes \cite{stewart_speech2spikes_2023} rely on first transforming audio recordings into spectrograms, a step not needed for already spectrotemporal visibility data.
In this work, we adapt the step-forward encoding algorithm to facilitate the direct readout of RFI flags by adding an additional small temporal element to the encoding method. This allows the output of the SNNs to be read as an RFI mask directly.
\section{Methods}
Due to the lack of preexisting information on detecting RFI with SNNs, we formulate the problem and define the encoding methods investigated.
Our principle contribution is formulating this problem so that the time axis included in the original data is unrolled into the SNN execution, translating what is typically considered a two-dimensional image segmentation problem into a multi-variate time-series segmentation problem. This reformulation allows us to interpret the output spike trains of an SNN as an RFI mask with minimal post-processing.
This section outlines our definition of RFI detection with SNNs, the parameters of our network design, and our approach toward hyper-parameter optimization.
\subsection{RFI Detection Formulation}
In radio astronomy, specialized hardware correlates raw signals from antennae, transforming the data from voltage information into complex-valued `visibilities' $V(\upsilon, T, b)$, which vary in frequency, time, and baseline (pair of antennae).
At this stage in the pipeline, RFI detection requires producing a boolean mask of `flags' $G(\upsilon, T, b)$ that vary in the same parameters but are binary-valued.
Previous supervised approaches to RFI detection formulate the problem as 
\begin{equation}\label{eq:originalformulation}
    \mathcal{L}_{sup} = 
    min_{\theta_n}\mathcal{H}(
        m_{\theta_n}(
            V(\upsilon, T, b)
        ), 
        G(\upsilon, T, b)
    ),
\end{equation}
where $\theta_n$ are the parameters of some classifier $m$ and $\mathcal{H}$ is an entropy-based similarity measure \cite{mesarcik_learning_2022}.
We exploit the time-varying nature of this information and, therefore, include additional spike encoding and decoding steps and push the similarity measure inwards, operating on each element of the resulting time-series.
\begin{equation}\label{eq:newformulation}
    \mathcal{L}_{sup} = 
    min_{\theta_n}(
        \Sigma_t^T
        \mathcal{H}(
            m_{\theta_n}(
                E(V(\upsilon, t, b))
            ), 
            F(G(\upsilon, t, b))
        )
    )
\end{equation}
where $E$ is an input encoding function and $F$ is an output encoding function.
The following six subsections describe different implementations of $E$ and $F$ and their accompanying comparison function, $\mathcal{H}$.
Figure \ref{fig:encodings} contains a raster plot for each encoding method presented with an exposure time of four where applicable, and the associated original spectrogram and mask in Figure \ref{fig:enc:original}.
\subsection{Encoding Methods}
\subsubsection{Latency}
Our latency encoding scheme is relatively straightforward. For a given exposure $E$, input pixel intensities are mapped inversely linearly from 0 to $E$, meaning high-intensity pixels spike almost immediately and low-intensity pixels progressively later. For output decoding, we interpret any spikes before the last exposure step as an RFI mask and no spikes or a spike in the final exposure as background. The following equation maps supervised masks:
\begin{equation}\label{eq:latency:mask}
    G(\upsilon, t, b)) = \begin{cases}
        0 & b = 1 \\
        E & otherwise
    \end{cases}
\end{equation}
The comparison function, $\mathcal{H}$, is the mean square spike time governed by the following function:
\begin{equation}\label{eq:latency:comparison}
    \mathcal{H}_{latency} = \Sigma_e^E\Sigma_\upsilon^\Upsilon(y_{\upsilon,e} - f_{\upsilon, e})^2.
\end{equation}
Figure \ref{fig:enc: latency} contains an example input raster plot.
\subsubsection{Rate}
Rate encoding is well-understood and widely used when processing image data in SNNs. For a given exposure $E$, we interpret pixel values as firing probabilities. We map supervised masks to firing rates of 0.8 for pixels labeled as containing RFI and 0.2 for pixels that do not. For output decoding, we interpret outputs with a firing rate greater than 0.75 as an RFI flag.

To map rate-encoding into a time-series segmentation problem, we treat each time-step in the original signal as a classification problem. The comparison function, $\mathcal{H}$, is an adjusted mean square error spike count loss, adapted from the implementation in snnTorch \cite{eshraghian_training_2023} to handle time segments with no positive labels present, and is governed by the following equation:
\begin{equation}
    \mathcal{H}_{rate} = \Sigma_e^E(y_{\upsilon,e} - f_{\upsilon, e})^2.
\end{equation}
Figure \ref{fig:enc:rate} contains an example input raster plot.
 \subsubsection{Delta-Modulation}
Delta-modulation encoding encodes inputs only on relative change. Our method utilizes the delta-modulation encoding method in snnTorch \cite{eshraghian_training_2023}, using both positive and negative polarity spikes. Output decoding is a more complex procedure.
Encoding regions of an RFI mask that can span as little as a single pixel requires spike polarity for `on' and `off' spikes. However, in our networks, neurons only emit positive spikes. Therefore, we double the frequency width of the output layer and move all negative polarity spikes to the lower region as positive spikes. 

The comparison function, $\mathcal{H}$, is the Huber loss function, which combines L1 and MSE loss functions, computing the squared MSE term if the error falls below a threshold and L1 loss otherwise. We employ this loss function to handle the sparsity of the output spike trains. The Huber loss as acting on spike-trains is:
\begin{equation}
    \mathcal{H}_{delta} = \begin{cases}
        \frac{1}{2}(y - f)^2 & |y - f| \leq \delta \\
        \delta (|y - f| - \frac{1}{2}\delta) & otherwise
    \end{cases}
\end{equation}

\subsubsection{Step-Forward}
Visibility data is spectrotemporal, so we tested encoding methods based on encoding audio data, the Step-Forward algorithm \cite{stewart_speech2spikes_2023}. This approach takes the step-wise difference between the signal at each time-step and a running cumulative sum, emitting a positive or negative spike when the difference crosses a threshold (in our case, $0.1$) upwards or downwards. Additionally, a cumulative sum of these differences is appended to the end of the frequency channels, doubling the width of the input. Output decoding and supervised mask encoding, covered in Equation \ref{eq:latency:mask}, are handled identically to latency encoding. The comparison function is also identical to latency encoding, covered in Equation \ref{eq:latency:comparison}. We provide three methods to adapt the original step-forward algorithm to the time-series segmentation problem outlined below; each method handles the required exposure parameter.
\paragraph{First}
This exposure mode presents spikes to the network only on the first exposure step for each time-step, with silence following.
Figure \ref{fig:enc:forwardstep:first} contains an example raster plot for this exposure method.
\paragraph{Direct}
This exposure mode presents spikes to the network at every exposure step for each time-step.
Figure \ref{fig:enc:forwardstep:direct} contains an example raster plot for this exposure method.
\paragraph{Latency}
This exposure mode presents spikes to the network latency encoded at exposure time $0$ where a spike is present and exposure time $E$ otherwise.
Figure \ref{fig:enc:forwardstep: latency} contains an example raster plot for this exposure method.
\begin{figure*}[htbp]
    \centering
    \begin{subfigure}{\columnwidth}
        \centering
        \includegraphics[width=0.9\textwidth, height=1.25in, keepaspectratio]{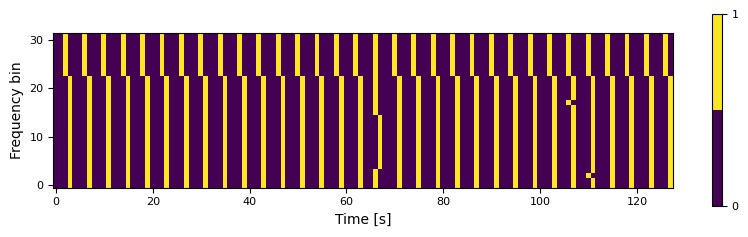}
        \caption{Latency encoded example. The background gradient is reflected across the entire spike-train, and the banding and blips are easily visible.}
        \label{fig:enc: latency}
    \end{subfigure}
    \begin{subfigure}{\columnwidth}
        \centering
        \includegraphics[width=0.9\textwidth, height=1.25in, keepaspectratio]{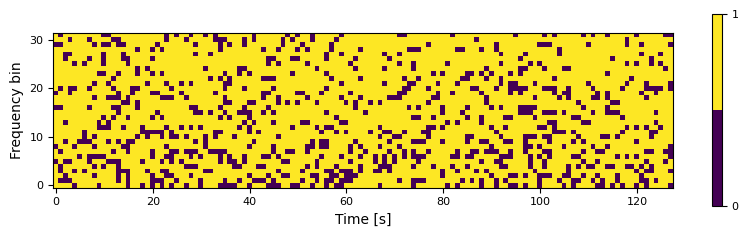}
        \caption{Rate encoded example. The features are not comprehensible in this encoding for such a small exposure value.}
        \label{fig:enc:rate}
    \end{subfigure}
    \begin{subfigure}{\columnwidth}
        \centering
        \includegraphics[width=0.9\textwidth, height=1.25in, keepaspectratio]{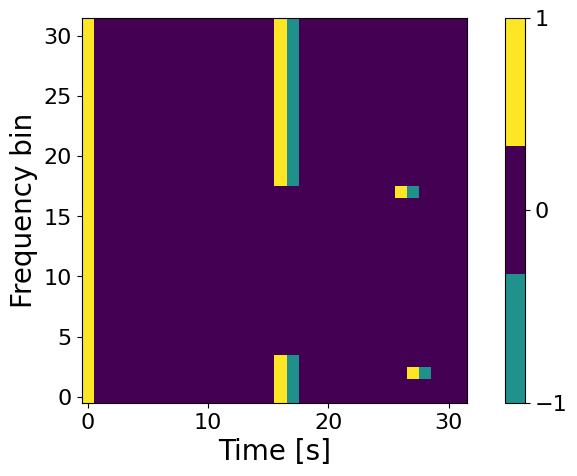}
        \caption{Delta encoded example. While most features are legible, and removing the gradient may be helpful, this encoding misses the banding below the background as a sub-threshold feature.}
        \label{fig:enc:delta}
    \end{subfigure}
    \begin{subfigure}{\columnwidth}
        \centering
        \includegraphics[width=0.9\textwidth, height=1.25in, keepaspectratio]{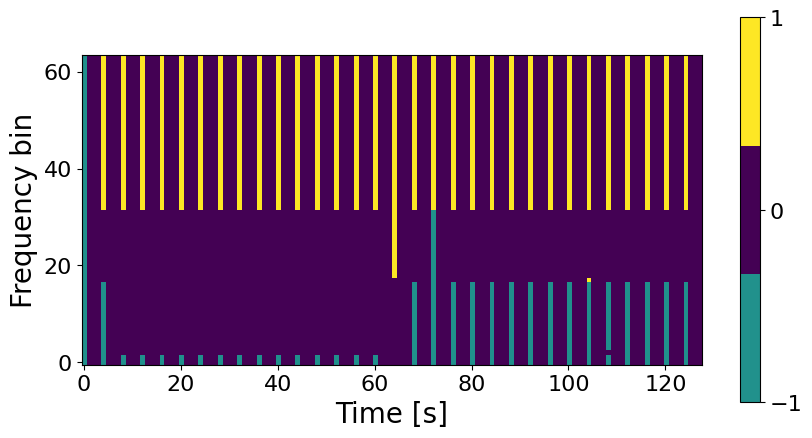}
        \caption{Step-Forward encoded example with `first' exposure mode. This encoding clearly shows the banding effect, however, the blips are not easily identifiable.}
        \label{fig:enc:forwardstep:first}
    \end{subfigure}
    \begin{subfigure}{\columnwidth}
        \centering
        \includegraphics[width=0.9\textwidth, height=1.25in, keepaspectratio]{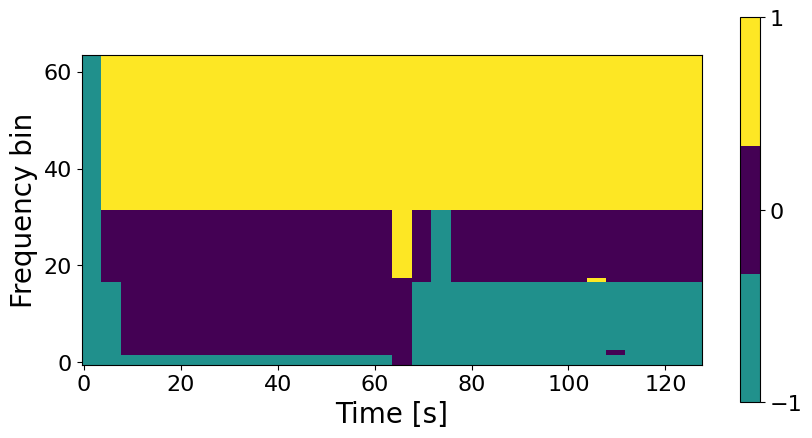}
        \caption{Step-Forward encoded example with `direct' exposure mode. This encoding is highly active however the blips are still hidden.}
        \label{fig:enc:forwardstep:direct}
    \end{subfigure}
    \begin{subfigure}{\columnwidth}
        \centering
        \includegraphics[width=0.9\textwidth, height=1.25in, keepaspectratio]{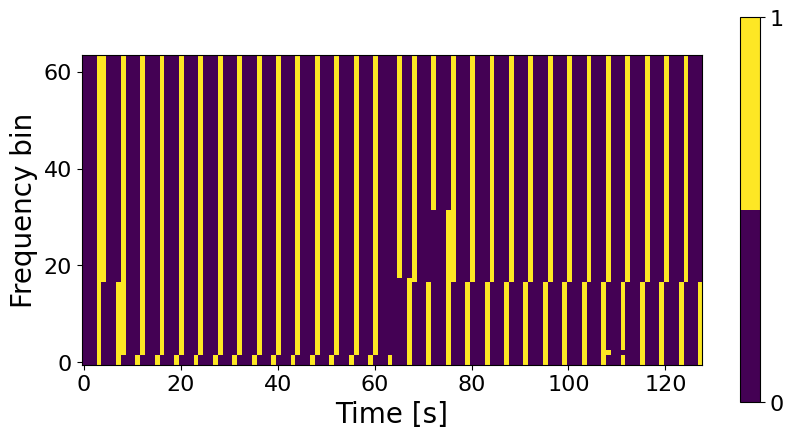}
        \caption{Step-Forward encoded example with `latency' exposure mode. This encoding, much like the original latency encoding, makes the background gradient, banded RFI, and blips easily identifiable.}
        \label{fig:enc:forwardstep: latency}
    \end{subfigure}
    \begin{subfigure}{\columnwidth}
        \centering
        \includegraphics[width=0.9\textwidth, height=1.25in, keepaspectratio]{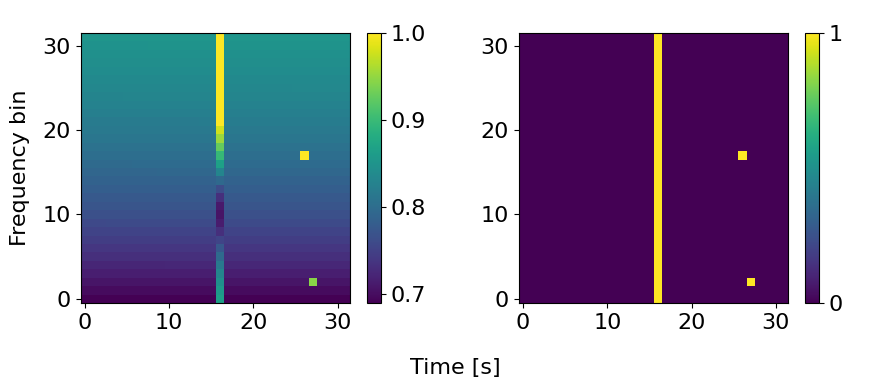}
        \caption{The original input spectrogram and associated RFI mask. This example gives a good indication of the type of challenge RFI detection presents. This example contains banded RFI present across all frequencies that is both above and below the background environment, two `blips' tightly localized in frequency and time, and a gradient in the background.}
        \label{fig:enc:original}
    \end{subfigure}
    \caption{Example spike raster plots for all explored encodings, the original spectrogram, and associated RFI mask. Encodings are exposed for four time-steps where applicable.}
\label{fig:encodings}
\end{figure*}
\subsection{Training Strategy}
We train a simple feedforward SNN on a dataset from the Hydrogen Epoch of Reionization Array (HERA) simulator \cite{mesarcik_learning_2022-1}. This dataset, described fully by Mesarcik et al. \cite{mesarcik_learning_2022}, comprises 420 training spectrograms, 140 test spectrograms, and 2.76\% RFI contamination by pixel count. Each spectrogram covers a 30-minute simulation integrating every 3.52 seconds (producing 512 time-steps) and 512 frequency channels from 105MHz to 195MHz. The RFI is synthetic but modeled from satellite communications (localized in frequency, covering all time), lightning (covering all frequencies, localized in time), ground communication (localized in frequency, partially localized in time), and impulse blips (localized in frequency and time). After encoding into spikes, we split each spectrogram further into $32 \times 32$ sub-spectrogram patches, allowing for a smaller test model and less intensive training requirements.
We chose to train a small two-layer fully connected network outlined in Table \ref{tab:networkarch}, avoiding complex structures and recurrent connects to simplify training and focus on the effectiveness of each encoding method specifically, even if this leads to lower overall task performance.
\begin{table}[htbp]
\centering
\caption{Spiking Neural Network Architecture, (*) indicates value for step-forward encodings and (\textdagger) for delta-modulation.}
\label{tab:networkarch}
\begin{tabular}{@{}lll@{}}
\toprule
\textbf{Layer} & \textbf{Input} & \textbf{Output} \\ \midrule
Dense 1        & 32 (64*)       & 128             \\
Dense 2        & 128            & 32 (64*\textdagger)        \\ \bottomrule
\end{tabular}
\end{table}

We trained our SNNs using surrogate-gradient backpropagation through time (BPTT) with snnTorch \cite{eshraghian_training_2023} using ATan as the surrogate gradient, combined with lightning\footnote{https://lightning.ai/} for data-driven parallelism when training. During training, we reset the internal network state after processing each sub-spectrogram, allowing temporal information to propagate through several time-steps (in the original spectrogram). We trained all models with Adam as the optimizer and utilized automatic learning rate scheduling with early stopping. All code is available online \cite{pritchard_pritchardnsnn-rfi-super_2024}.
\subsection{Hyper-Parameter Tuning}
We employ Optuna \cite{akiba_optuna_2019} for hyper-parameter optimization, specifically the tree-structured Parzan estimation algorithm, to simultaneously conduct multi-variate optimization for Area Under the Receiver Operating Characteristic curve (AUROC), Area Under the Precision-Recall Curve (AUPRC), Accuracy, and F1-Score.
AUROC evaluates the ratio of True Positive Rate (TPR) and False Positive Rate (FPR) across all possible classification thresholds.
AUPRC gives the ratio of precision and recall across several thresholds, in this case referring to the proportion of correctly classified RFI over all RFI predictions. Recall in this case is the TPR.
Accuracy is the per-pixel output accuracy. For this synthetic radio astronomy dataset, we expect accuracy to be very high as true RFI flags are very sparse; a completely silent network would still be more than 90\% accurate.
F1-Score is the harmonic mean of precision and recall at a given threshold.
The sparsity of RFI in this dataset means we generally expect high AUROC and high accuracy values with lower AUPRC and F1 scores. A good-performing model needs to be silent the majority of the time while providing confident RFI detections, hence our inclusion of both raw accuracy and f1-scores.
Table \ref{tab:hyperparam-range} contains the bounds of each variable included in the optimization for all encoding methods.
We conduct 100 trials for each encoding method.
\begin{table}[htbp]
\centering
\caption{Parameter ranges for attributes included in hyper-parameter searches. Beta defines the excitability of LiF neurons within the network, and exposure is a parameter for the encoding method used in all but the delta-modulation method.}
\label{tab:hyperparam-range}
\begin{tabular}{@{}ll@{}}
\toprule
Attribute  & Parameter Range \\ \midrule
Batch Size & 16 - 128        \\
Epochs     & 5 - 100         \\
Beta       & 0.5 - 0.99      \\
Exposure   & 1 - 64          \\ \bottomrule
\end{tabular}
\end{table}

For each encoding method, we take the trials with the best score in any metric into a final set and subsequently compare these trials between all encoding techniques to determine the overall best encoding method and set of parameters. Balancing the rate of neuron-state decay (beta) with the exposure of each timestep in the spectrogram is challenging; more exposure increases the fidelity of the input data but dilutes the information propagating over real time. Moreover, a changing exposure changes the input data shape hence why we allow Optuna to select batch size. We discuss the result of this process in Section \ref{sec:results} as they comprise an integral part of our investigation.
Finally, we perform fifty repeat trials for the best parameters in each encoding method to confirm their performance.
\section{Results}\label{sec:results}
\subsection{Encoding Comparison}
We present the results of our hyper-parameter searches in Table  \ref{tab:results:hyperparams}. No prior investigation has been conducted into encoding radio astronomy visibility data into SNNs for RFI detection, and as such, these results are of interest. For each method, we present the number of trials conducted, the best candidate set (an artifact of performing a multi-variate optimization), and the best trials whose performance in at least one metric (accuracy, AUROC, AUPRC, or F1-Score) is best. We ran fifty trials for the best parameters for each encoding method and present the results in Table \ref{tab:results}.
\begin{table*}[htbp]
\centering
\caption{Hyper-Parameter search using Optuna multi-variate optimization. We show the top three to five trials per encoding method, as Optuna selects during its multi-objective optimization routine. A range of beta and exposure values produce decent results in all cases. Trials with smaller exposure tend to score higher in raw accuracy. Larger exposures tend to combine with larger beta values to allow information to propagate throughout the entire spectrogram sample. The sparsity of RFI in radio astronomy spectrograms means models tend to perform better in accuracy and AUROC than in AUPRC and F1-Scores, where performance focuses on the positive class. The best trial for each method is bolded, and the best SNN encoding method is underlined. A standard ANN narrowly produced the highest overall performing hyper-parameters in a single trial, producing top AUPRC and F1-Scores.}
\label{tab:results:hyperparams}
\begin{tabular}{@{}cccccccccc@{}}
\toprule
Encoding Method          & Accuracy                  & AUROC          & AUPRC          & F1-Score       & Batch Size   & Epochs      & Beta           & Exposure    \\ \midrule
Delta-Modulation         &                &                &                &                &                &              &             &                &             \\
                         & \textbf{0.983} & \textbf{0.846} & \textbf{0.697} & \textbf{0.676} & \textbf{48}  & \textbf{80} & \textbf{0.504} & - \\
                         & 0.981                   & 0.849          & 0.668          & 0.620          & 19           & 37          & 0.555          & -          \\
                         & 0.978                   & 0.848          & 0.655          & 0.632          & 62           & 59          & 0.564          & -          \\
Latency                  &                &                &                &                &                &              &             &                &             \\
                         & \textbf{0.967}   & \textbf{0.846} & \textbf{0.674} & \textbf{0.668} & \textbf{36}  & \textbf{44} & \textbf{0.727} & \textbf{6}  \\
                         & 0.958                     & 0.640          & 0.296          & 0.286          & 127          & 29          & 0.720          & 2           \\
                         & 0.960                   & 0.862          & 0.650          & 0.633          & 27           & 34          & 0.842          & 13          \\
Step-Forward-Direct      &                &                &                &                &                &              &             &                &             \\
                         & \textbf{0.988}  & \textbf{0.938} & \textbf{0.789} & \textbf{0.768} & \textbf{31}  & \textbf{43} & \textbf{0.920} & \textbf{50} \\
                         & 0.988                   & 0.929          & 0.740          & 0.697          & 27           & 14          & 0.630          & 17          \\
                         & 0.983                  & 0.947          & 0.674          & 0.575          & 42           & 77          & 0.836          & 9           \\
                         & 0.990                   & 0.796          & 0.694          & 0.645          & 18           & 42          & 0.980          & 16          \\
{\ul Step-Forward-First} &                &                &                &                &                &              &             &                &             \\
                         & \textbf{0.990}  & \textbf{0.892} & \textbf{0.857} & \textbf{0.844} & \textbf{79}  & \textbf{98} & \textbf{0.949} & \textbf{41} \\
                         & 0.989                   & 0.978          & 0.773          & 0.747          & 22           & 38          & 0.984          & 41          \\
                         & 0.995                   & 0.914          & 0.730          & 0.686          & 19           & 60          & 0.666          & 2           \\
                         & 0.982                 & 0.939          & 0.667          & 0.566          & 30           & 14          & 0.960          & 5           \\
Step-Forward-Latency     &                &                &                &                &                &              &             &                &             \\
                         & \textbf{0.966}  & \textbf{0.900} & \textbf{0.757} & \textbf{0.731} & \textbf{54}  & \textbf{83} & \textbf{0.921} & \textbf{22} \\
                         & 0.966                   & 0.0875         & 0.719          & 0.697          & 112          & 40          & 0.858          & 20          \\
                         & 0.972                  & 0.500          & 0.514          & 0.052          & 106          & 96          & 0.657          & 1           \\
                         & 0.958                  & 0.907          & 0.616          & 0.478          & 18           & 79          & 0.579          & 3           \\
                         & 0.968                  & 0.897          & 0.759          & 0.727          & 40           & 93          & 0.935          & 18          \\
Rate                     &                &                &                &                &                &              &             &                &             \\
                         & \textbf{0.972}  & \textbf{0.500} & \textbf{0.514} & \textbf{0.052} & \textbf{107} & \textbf{50} & \textbf{0.599} & \textbf{1}  \\
                         & 0.917                   & 0.511          & 0.070          & 0.056          & 114          & 36          & 0.781          & 4           \\
                         & 0.028                  & 0.500          & 0.514          & 0.052          & 119          & 80          & 0.732          & 32          \\
ANN                      &                &                &                &                &                &              &             &                &             \\
                         & \textbf{0.989} & \textbf{0.901} & \textbf{0.870} & \textbf{0.845} & \textbf{44}  & \textbf{29} & \textbf{-}     & \textbf{-}  \\
                         & 0.978               & 0.956          & 0.782          & 0.711          & 26           & 56           & -              & -           \\
                         & 0.982                 & 0.953          & 0.827          & 0.793          & 49          & 75          & -              & - \\ \bottomrule          
\end{tabular}
\end{table*}
\begin{table*}[htbp]
\centering
\caption{Encoding method performance using final hyper-parameters. Each metric is listed as mean and standard deviation. The best scores are bolded. Latency encoding offers superior performance in all metrics and with less variability than all but rate-encoded methods. ANN performance exhibits the highest overall variability.}
\label{tab:results}
\begin{tabular}{@{}cccccccccc@{}}
\toprule
Encoding Method      & \multicolumn{2}{c}{Accuracy}             & \multicolumn{2}{c}{AUROC}       & \multicolumn{2}{c}{AUPRC}       & \multicolumn{2}{c}{F1}          & Num Trials \\ \midrule
Delta                & 0.963          & 0.037          & 0.840          & 0.013          & 0.630          & 0.075          & 0.571          & 0.126          & 50         \\
\textbf{Latency}     & \textbf{0.988} & \textbf{0.004} & \textbf{0.929} & 0.004          & \textbf{0.785} & 0.035          & \textbf{0.761} & 0.050          & 50         \\
Step-Forward-Direct  & 0.976          & 0.015          & 0.856          & 0.022          & 0.681          & 0.071          & 0.643          & 0.105          & 50         \\
Step-Forward-First   & 0.975          & 0.015          & 0.859          & 0.021          & 0.679          & 0.074          & 0.637          & 0.109          & 50         \\
Step-Forward-Latency & 0.972          & 0.015          & 0.863          & 0.021          & 0.662          & 0.073          & 0.611          & 0.108          & 50         \\
Rate                 & 0.943          & 0.009          & 0.513          & \textbf{0.003} & 0.066          & \textbf{0.006} & 0.059          & \textbf{0.002} & 50         \\
ANN                  & 0.816          & 0.181          & 0.846          & 0.081          & 0.595          & 0.110 & 0.368  & 0.223 & 50         \\ \bottomrule
\end{tabular}
\end{table*}

The best SNN encoding scheme based on the hyper-parameter search is the step-forward-first method, with an accuracy of 99\%, AUROC of $0.89$, AUPRC of $0.857$, and F1-Score of $0.844$. The step-forward-direct and step-forward-latency methods also perform strongly. Combining the delta-modulated changes and sustained value helps encode visibility data, which is similarly effective to the speech2spikes encoding method, albeit demonstrated on a classification task. Latency and Delta-modulation encoding perform similarly well to each other. Rate encoding is deeply non-competitive, with an accuracy of 15\% and an F1-score of $0.064$.
We conducted a hyper-parameter search and evaluation with an ANN of the identical two-layer architecture of the SNNs. We found that the single-trial performance in the hyper-parameter search narrowly outperformed all SNN methods by providing top AUPRC and F1-Scores.

The repeat trials, however, tell a different story. Latency encoding outperforms all other methods in terms of raw performance and lower variability. We conclude that latency encoding is the overall best-performing SNN encoding method for this task. Figure \ref{fig:example} contains example output RFI masks for a latency-encoded model. The trained SNN manages to detect major RFI features but, unsurprisingly, produces some misclassifications on individual pixels.
Moreover, ANN performance is surprisingly much more variable and performs worse on average than all SNN methods except rate-encoding.
\begin{figure}[htbp]
    \centering
    \begin{subfigure}{\columnwidth}
        \centering
        \includegraphics[width=\textwidth, height=1.5in, keepaspectratio]{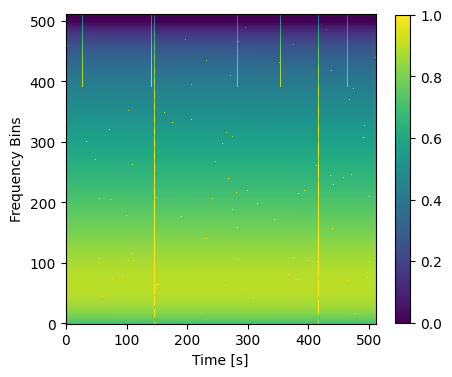}
        \caption{Original spectrogram.}
        \label{fig:res:orig}
    \end{subfigure}
    \begin{subfigure}{\columnwidth}
        \centering
        \includegraphics[width=\textwidth, height=1.5in, keepaspectratio]{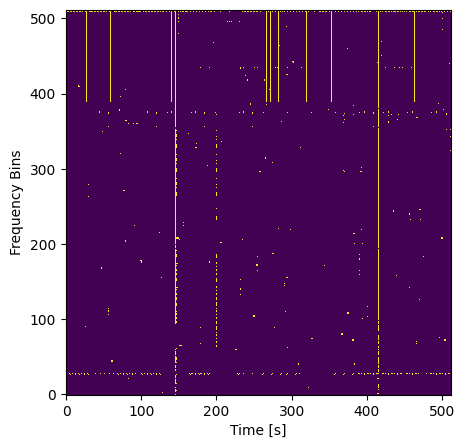}
        \caption{Latency-based inference.}
        \label{fig:res:inf}
    \end{subfigure}
    \begin{subfigure}{\columnwidth}
        \centering
        \includegraphics[width=\textwidth, height=1.5in, keepaspectratio]{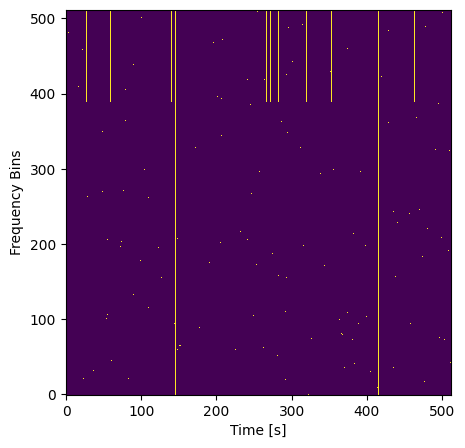}
        \caption{Ground-truth RFI mask.}
        \label{fig:res:mask}
    \end{subfigure}
    \caption{Example HERA Spectrogram, the output of a latency-encoded trained SNN, and the original RFI mask.}
\label{fig:example}
\end{figure}
\section{Discussion}
The current state-of-the-art performance for any RFI detection method on this HERA dataset achieves an AUROC of $0.983$, AUPRC of $0.940$, and F1-Score of $0.939$ \cite{mesarcik_learning_2022, pritchard_rfi_2024}. The best performance of any encoding method tested achieves an AUROC of $0.929$, AUPRC of $0.785$, and F1-score of $0.761$. Our method is surprisingly competitive with other methods tested on this dataset, which is promising given the simplicity and size of the SNN tested in this work.

Re-formulating the task of RFI detection from an image-domain semantic segmentation problem to a time-series segmentation problem permits the use of simple SNN networks as the input width of the network only includes frequency channels, leaving the spiking neurons to handle the temporal information instead of including two dimensions of information as is typical when treating the problem as an image-segmentation task. Our hyper-parameter optimization shows that the encoding methods that maximally leverage temporal information, in this case, latency and step-forward encoding, achieve the best overall performance under this formulation. We additionally show that rate-encoding, typically employed when converting static image data to a spike-coded modality or ANN2SNN conversion, performs poorly in this formulation.
In combination with the ANN model's poor and highly variable performance, it is possible that the problem of RFI detection, as formulated as a time-series segmentation problem, may be particularly well suited to SNN evaluation.
\subsection{Limitations}
While we endeavor to cover various encoding techniques, our investigation is not exhaustive.
Moreover, training for small segments of a spectrogram, limited in frequency and time, may limit the accuracy possible in this formulation of RFI detection. Ideally, such an RFI scheme would operate on a continual streamed input of visibility information for an entire observation.
The HERA dataset tested is synthetic, and we know most methods will likely overfit this simulated data. Therefore, performing performance testing on real astronomical observations is desirable to gain a more realistic understanding of our method's performance.
Finally, a supervised approach requires high-quality supervised labels, which are difficult to find in practice \cite{mesarcik_learning_2022}. 
\subsection{Future Direction}
This work implements a method for RFI detection that achieves excellent performance, considering the small size and simplicity of the SNNs tested.
However, a performance gap exists between the state-of-the-art performance and the dataset tested today. An obvious extension to this work includes expanding the network architectures. Expansions to network architectures include larger, fully connected networks, adding recurrence relationships, and incorporating some spectral information through convolution. Moreover, conducting network architecture searches and evolutionary optimization may reveal interesting patterns.
Testing the performance of larger SNNs trained on patches of spectrograms encoded in two dimensions, based on the original RFI detection formulation in Equation \ref{eq:originalformulation} will also determine if there is any efficiency benefit to our time-series formulation specifically.
Expanding the datasets tested to include real observational data is a second obvious extension.
It would also be interesting to test other learning rules, such as supervised membrane-based backpropagation, online unsupervised bio-inspired learning rules, or reservoir computing.
Finally, mapping an SNN-based RFI detection approach to available neuromorphic hardware would test a possible operational improvement in energy usage or computational performance.
\subsection{Conclusion}
We formulate RFI detection as a time-series segmentation problem suitable for direct SNN inference and evaluate the accuracy of six encoding methods. We find latency encoding the most performant on the synthetic HERA dataset tested here. Although there is a gap to the state-of-the-art non-SNN methods for RFI detection, the small size and simplicity of the SNN networks tested here strongly indicate that SNNs have a future in RFI detection both as a potential solution to RFI detection in contemporary radio telescopes and as a data-intensive benchmark for neuromorphic computing platforms.
\bibliographystyle{IEEEtran}
\bibliography{IEEEabrv,SupervisedSNN-ICONS24}

\begin{thebibliography}{10}
\providecommand{\url}[1]{#1}
\csname url@samestyle\endcsname
\providecommand{\newblock}{\relax}
\providecommand{\bibinfo}[2]{#2}
\providecommand{\BIBentrySTDinterwordspacing}{\spaceskip=0pt\relax}
\providecommand{\BIBentryALTinterwordstretchfactor}{4}
\providecommand{\BIBentryALTinterwordspacing}{\spaceskip=\fontdimen2\font plus
\BIBentryALTinterwordstretchfactor\fontdimen3\font minus \fontdimen4\font\relax}
\providecommand{\BIBforeignlanguage}[2]{{%
\expandafter\ifx\csname l@#1\endcsname\relax
\typeout{** WARNING: IEEEtran.bst: No hyphenation pattern has been}%
\typeout{** loaded for the language `#1'. Using the pattern for}%
\typeout{** the default language instead.}%
\else
\language=\csname l@#1\endcsname
\fi
#2}}
\providecommand{\BIBdecl}{\relax}
\BIBdecl

\bibitem{noauthor_report_2022}
\BIBentryALTinterwordspacing
``\BIBforeignlanguage{EN}{Report of the {Committee} on the {Peaceful} {Uses} of {Outer} {Space}},'' United Nations Office for Outer Space Affairs, Tech. Rep.~65, Aug. 2022. [Online]. Available: \url{https://www.unoosa.org/oosa/oosadoc/data/documents/2022/a/a7720\_0.html}
\BIBentrySTDinterwordspacing

\bibitem{offringa_aoflagger_2010}
A.~Offringa, ``{AOFlagger}: {RFI} {Software},'' \emph{Astrophysics Source Code Library}, pp. ascl--1010, 2010.

\bibitem{akeret_radio_2017}
\BIBentryALTinterwordspacing
J.~Akeret, C.~Chang, A.~Lucchi, and A.~Refregier, ``\BIBforeignlanguage{en}{Radio frequency interference mitigation using deep convolutional neural networks},'' \emph{\BIBforeignlanguage{en}{Astronomy and Computing}}, vol.~18, pp. 35--39, Jan. 2017. [Online]. Available: \url{https://www.sciencedirect.com/science/article/pii/S2213133716301056}
\BIBentrySTDinterwordspacing

\bibitem{kasabov_evolving_2016}
\BIBentryALTinterwordspacing
N.~Kasabov, N.~M. Scott, E.~Tu, S.~Marks, N.~Sengupta, E.~Capecci, M.~Othman, M.~G. Doborjeh, N.~Murli, R.~Hartono, J.~I. Espinosa-Ramos, L.~Zhou, F.~B. Alvi, G.~Wang, D.~Taylor, V.~Feigin, S.~Gulyaev, M.~Mahmoud, Z.-G. Hou, and J.~Yang, ``Evolving spatio-temporal data machines based on the {NeuCube} neuromorphic framework: {Design} methodology and selected applications,'' \emph{Neural Networks}, vol.~78, pp. 1--14, Jun. 2016. [Online]. Available: \url{https://www.sciencedirect.com/science/article/pii/S0893608015001860}
\BIBentrySTDinterwordspacing

\bibitem{scott_evolving_2015}
\BIBentryALTinterwordspacing
N.~M. Scott, ``\BIBforeignlanguage{en}{Evolving {Spiking} {Neural} {Networks} for {Spatio}- and {Spectro}- {Temporal} {Data} {Analysis}: {Models}, {Implementations}, {Applications}},'' Ph.D. dissertation, Auckland University of Technology, 2015. [Online]. Available: \url{https://hdl.handle.net/10292/10601}
\BIBentrySTDinterwordspacing

\bibitem{pritchard_rfi_2024}
\BIBentryALTinterwordspacing
N.~J. Pritchard, A.~Wicenec, M.~Bennamoun, and R.~Dodson, ``\BIBforeignlanguage{en}{{RFI} {Detection} with {Spiking} {Neural} {Networks}},'' \emph{\BIBforeignlanguage{en}{Publications of the Astronomical Society of Australia}}, pp. 1--11, Apr. 2024. [Online]. Available: \url{https://www.cambridge.org/core/journals/publications-of-the-astronomical-society-of-australia/article/rfi-detection-with-spiking-neural-networks/994EA6FDFA18B7799D7F0552264111E6}
\BIBentrySTDinterwordspacing

\bibitem{yang_deep_2020}
\BIBentryALTinterwordspacing
Z.~Yang, C.~Yu, J.~Xiao, and B.~Zhang, ``Deep residual detection of radio frequency interference for {FAST},'' \emph{Monthly Notices of the Royal Astronomical Society}, vol. 492, no.~1, pp. 1421--1431, Feb. 2020. [Online]. Available: \url{https://doi.org/10.1093/mnras/stz3521}
\BIBentrySTDinterwordspacing

\bibitem{wolfaardt_machine_2016}
\BIBentryALTinterwordspacing
C.~J. Wolfaardt, ``\BIBforeignlanguage{en}{Machine learning approach to radio frequency interference({RFI}) classification in {Radio} {Astronomy}},'' Thesis, Stellenbosch : Stellenbosch University, Mar. 2016, accepted: 2016-03-09T14:22:12Z. [Online]. Available: \url{https://scholar.sun.ac.za:443/handle/10019.1/98464}
\BIBentrySTDinterwordspacing

\bibitem{mesarcik_learning_2022}
\BIBentryALTinterwordspacing
M.~Mesarcik, A.-J. Boonstra, E.~Ranguelova, and R.~V. van Nieuwpoort, ``Learning to detect radio frequency interference in radio astronomy without seeing it,'' \emph{Monthly Notices of the Royal Astronomical Society}, vol. 516, no.~4, pp. 5367--5378, Nov. 2022. [Online]. Available: \url{https://doi.org/10.1093/mnras/stac2503}
\BIBentrySTDinterwordspacing

\bibitem{dutoit_comparison_2024}
\BIBentryALTinterwordspacing
C.~D. Du Toit, T.~L. Grobler, and D.~J. Ludick, ``A comparison framework for deep learning {RFI} detection algorithms,'' \emph{Monthly Notices of the Royal Astronomical Society}, vol. 530, no.~1, pp. 613--629, May 2024. [Online]. Available: \url{https://doi.org/10.1093/mnras/stae892}
\BIBentrySTDinterwordspacing

\bibitem{mesarcik_learning_2022-1}
\BIBentryALTinterwordspacing
M.~Mesarcik, A.-J. Boonstra, R.~van Nieuwpoort, and Ranguelova, ``Learning to detect {RFI} in radio astronomy without seeing it,'' Jun. 2022. [Online]. Available: \url{https://zenodo.org/records/6724065}
\BIBentrySTDinterwordspacing

\bibitem{yi_learning_2023}
\BIBentryALTinterwordspacing
Z.~Yi, J.~Lian, Q.~Liu, H.~Zhu, D.~Liang, and J.~Liu, ``\BIBforeignlanguage{en}{Learning {Rules} in {Spiking} {Neural} {Networks}: {A} {Survey}},'' \emph{\BIBforeignlanguage{en}{Neurocomputing}}, Feb. 2023. [Online]. Available: \url{https://www.sciencedirect.com/science/article/pii/S0925231223001662}
\BIBentrySTDinterwordspacing

\bibitem{stewart_speech2spikes_2023}
\BIBentryALTinterwordspacing
K.~M. Stewart, T.~Shea, N.~Pacik-Nelson, E.~Gallo, and A.~Danielescu, ``{Speech2Spikes}: {Efficient} {Audio} {Encoding} {Pipeline} for {Real}-time {Neuromorphic} {Systems},'' in \emph{Proceedings of the 2023 {Annual} {Neuro}-{Inspired} {Computational} {Elements} {Conference}}, ser. {NICE} '23.\hskip 1em plus 0.5em minus 0.4em\relax New York, NY, USA: Association for Computing Machinery, Apr. 2023, pp. 71--78. [Online]. Available: \url{https://dl.acm.org/doi/10.1145/3584954.3584995}
\BIBentrySTDinterwordspacing

\bibitem{eshraghian_training_2023}
J.~K. Eshraghian, M.~Ward, E.~Neftci, X.~Wang, G.~Lenz, G.~Dwivedi, M.~Bennamoun, D.~S. Jeong, and W.~D. Lu, ``Training spiking neural networks using lessons from deep learning,'' \emph{Proceedings of the IEEE}, vol. 111, no.~9, pp. 1016--1054, 2023.

\bibitem{pritchard_pritchardnsnn-rfi-super_2024}
\BIBentryALTinterwordspacing
N.~Pritchard, ``pritchardn/{SNN}-{RFI}-{SUPER}: {Initial} {Release},'' Apr. 2024. [Online]. Available: \url{https://zenodo.org/records/10929492}
\BIBentrySTDinterwordspacing

\bibitem{akiba_optuna_2019}
T.~Akiba, S.~Sano, T.~Yanase, T.~Ohta, and M.~Koyama, ``Optuna: {A} {Next}-generation {Hyperparameter} {Optimization} {Framework},'' in \emph{Proceedings of the 25th {ACM} {SIGKDD} {International} {Conference} on {Knowledge} {Discovery} and {Data} {Mining}}, 2019.

\end{thebibliography}
\end{document}